\documentclass[runningheads]{llncs}

\usepackage{eccv}

\usepackage{eccvabbrv}

\usepackage{amsmath}
\usepackage{amssymb}
\usepackage{colortbl}
\usepackage{xcolor}
\usepackage{caption}
\usepackage{subcaption}
\usepackage{multirow}
\usepackage{todonotes}

\usepackage{graphicx}
\usepackage{booktabs}
\usepackage{tabularx} 

\usepackage[accsupp]{axessibility}  

\usepackage{orcidlink}

\begin{document}

\title{PEAVS: Perceptual Evaluation of Audio-Visual Synchrony Grounded in Viewers' Opinion Scores} 

\titlerunning{PEAVS}

\author{Lucas Goncalves\thanks{Work done during internship at Amazon.}\inst{1} \and
Prashant Mathur\inst{2} \and
Chandrashekhar Lavania\inst{2} \and
Metehan Cekic\inst{2} \and
Marcello Federico\inst{2} \and
Kyu J. Han\inst{2}
}

\authorrunning{Goncalves et al. 2024}

\institute{The University of Texas at Dallas 
\and
AWS AI Labs\\
\email{pramathu@amazon.com}}

\maketitle

\begin{abstract}
Recent advancements in audio-visual generative modeling have been propelled by progress in deep learning and the availability of data-rich benchmarks. However, the growth is not attributed solely to models and benchmarks. Universally accepted evaluation metrics also play an important role in advancing the field. While there are many metrics available to evaluate audio and visual content separately, there is a lack of metrics that offer a quantitative and interpretable measure of audio-visual synchronization for videos `in the wild'. To address this gap, we first created a large scale human annotated dataset ({100+} hrs) representing nine types of synchronization errors in audio-visual content and how human perceive them. We then developed a PEAVS (Perceptual Evaluation of Audio-Visual Synchrony) score, a novel automatic metric with a 5-point scale that evaluates the quality of audio-visual synchronization. We validate PEAVS using a newly generated dataset, achieving a Pearson correlation of 0.79 at the set level and 0.54 at the clip level when compared to human labels. In our experiments, we observe a relative gain 50\% over a natural extension of Fr\'echet based metrics for Audio-Visual synchrony, confirming PEAVS’ efficacy in objectively modeling subjective perceptions of audio-visual synchronization for videos `in the wild'.
\end{abstract}

\section{Introduction}
\label{sec:intro}

Audio-visual (AV) generative modeling has made rapid progress in recent years due to advancements in deep learning models \cite{vaswaniattention} and the availability of large-scale datasets like VGGSound \cite{chen2020vggsound} and AudioSet \cite{audioset}. However, evaluation of these models remains an open challenge.
Existing automatic metrics often focus on specific aspects, such as image quality~\cite{unterthiner2019accurate,hessel_2021} or audio fidelity~\cite{Kilgour_2019,roux_2018,chinen_2020}, lacking a satisfactory measure for assessing audio-visual synchronization of `in the wild' videos.
In fact, lack of well-established metrics for AV synchrony issues can already be observed in literature. For instance, recent works in the realm of audio-visual generation, like MM-Diffusion \cite{ruan2023mmdiffusion}, had to solely rely on single-modality metrics to evaluate their outputs. The work on Diff-Foley \cite{luo2023difffoley} had to train their own audio-visual synchrony classifier.
More advanced evaluation metrics are needed to provide a holistic assessment of audio-visual coherence. The development of such metrics is essential to accurately judge model performance, identify failure cases, and drive further progress in this evolving field.




Recent studies such as AV Synchrony Transformer (AVST)~\cite{chen2021audiovisual}, SparseSync~\cite{iashin2022sparse} and Diff-Foley~\cite{luo2023difffoley} have defined the synchronization problem between audio-visual modalities primarily as an issue of temporal offset determination. However, this perspective addresses only one aspect of synchronization challenges. In our work, we extend the scope to a broader range of synchronization issues, encompassing audio/visual speed variations, intermittent muting, fragment shuffling, AV flickering, and temporal shifts. We also delve into developing a perceptual, automatic metric that aligns with human judgment. To achieve this, we initiated a large scale annotation study, gathering human assessments on audio-visual synchrony across over {100} hours of both original and distorted content, featuring the aforementioned synchronization issues.

In this endeavor, our objective is to create a metric that accomplishes the following: (a) attains a high correlation with human assessments of synchronization issues, (b) offers interpretable scoring for end users, and (c) operates in a reference-free manner, meaning it does not rely on ground truth for quality prediction. The key innovation in our approach lies in the development of an interpretable scale based on detailed perceptual guidelines provided to human raters. By optimizing for concordance with human-annotated scores during training, our metric produces scores directly aligned with predefined levels of audio-visual (a)synchrony. This emphasis on interpretability and alignment with human perception addresses the need for a more insightful evaluation of AV synchronization.


\noindent Our main contributions are as follows:
\begin{itemize}

   \item We present a new Audio-Visual Synchrony human perception (AVS) benchmark data set with over 120K annotations and over {100+} hrs of content. Each video in this benchmark is annotated by three different annotators and we see an agreement of 0.71 Krippendorff's alpha~\cite{krippendorff04}.\looseness-1
   \item We propose a new audio-visual synchrony evaluation metric - PEAVS - that is reference-free and has interpretable scoring on a scale of 1 to 5.
   \item PEAVS shows high correlation of 0.79 with human judgements on the benchmark data, significantly outperforming a Fréchet distance based AV synchrony metric.
   
 \end{itemize} 


\section{Related Works}

\paragraph{Unimodal Metrics:} Several factors have fueled progress in single modality generation, including advancement in deep learning model architectures and training/optimization methodology, increased computational capabilities, a rich availability of multimodal data, and the development of consistent metrics. Established metrics like Inception Score (ISc) \cite{salimans_2016}, Fréchet Inception Distance (FID) \cite{Heusel_2017}, Fréchet Audio Distance (FAD) \cite{Kilgour_2019}, Fréchet Video Distance (FVD) \cite{unterthiner2019accurate}, inter alia, are frequently employed to gauge the quality of the content generated by unimodal generative models, providing a standard for comparison. However, these metrics are reliant only on one modality - either audio \cite{Kilgour_2019,roux_2018,Vincent_2006,luo_2018,Rix_2001,taal_2011,chinen_2020}, video \cite{unterthiner2019accurate,hessel_2021,zhang_2018,Wang_2002,wang_2003,Korhonen_2012,bińkowski_2021}, or just applied to images \cite{salimans_2016, Heusel_2017}. In the evaluation of generated multimodal content, ensuring the quality of individual modalities is only half the battle. It is crucial to evaluate the synchronization between audio and visual elements, especially in the context of audio-visual generation.\looseness-1

\paragraph{Multimodal Metrics:} For AV synchrony evaluation, it is important to determine the misalignment between audio and visual inputs. In the past, researchers have explored handling synchrony by statistical approaches and specifically designed features to pinpoint where the sound comes from, like locating the source of a voice in a scene \cite{Hershey_2020}. However, with the rise of deep learning recent studies are harnessing distinct connections between audio and visuals, like those in human speech \cite{Chung16a}, {audio-visual correspondence to check for matching content\cite{Arandjelovi_2017}}, or musical performances \cite{arandjelović2018objects}. Some recent research studies, such as those by Chen \textit{et al.} \cite{chen2021audiovisual} and Iashin \textit{et al.} \cite{iashin2022sparse}, 
propose methods to determine audio-visual synchronization in a natural diverse video environment. However, their focus is solely on tackling the synchronization problem arising from delays in audio or video streams, specifically addressing differences in timing (offsets). In these existing approaches, the model is trained with the objective of either 1) detecting synchronization errors as a binary classification or 2) predicting the precise offset values through regression. In contrast, our aim is to have the model replicate the way humans perceive synchronization issues across a range of distortion types (see more in Sec~\ref{sec:avdistortions}).\looseness-1

\paragraph{Metrics for ``in the wild'' videos:} The assessment of multimodal content has been extensively investigated within the computer vision community however most of the work has been tailored to lip-synchrony for instance, SyncNet~\cite{Chung16a}  quantifies lip-sync errors by comparing mouth movements to the corresponding speech; AlignNet~\cite{jianren20alignnet} focuses on aligning video content with ground truth audio, and predicts frame-wise offsets.  
In this work, we focus on evaluation of AV synchrony for ``in the wild'' videos which apart from Chen \textit{et al.} \cite{chen2021audiovisual}, Iashin \textit{et al.} \cite{iashin2022sparse} has been explored in Diff-Foley~\cite{luo2023difffoley} where the authors train their own alignment classifier to detect synchrony. However, none of these works directly target modeling of human perception.

\paragraph{Perceptual Metrics:} Previous studies on audio metrics \cite{manocha_differentiable_2020,manocha2021} demonstrated that a learned metric trained on data collected via crowd-sourced human annotations, particularly on just noticeably different samples, correlates well with mean opinion scores.
For video modality, Wang et. al. \cite{wang2021perceptual} designed a perceptual metric --- learnt over a large scale human annotated data --- designed to distinguish user generated video content in different dimensions such as semantic content, technical quality, and compression level. Our work in contrast centers on the synchronization between audio and video than solely on video or audio quality.





\begin{table*}
\begin{tabularx}{\textwidth}{l | c | X | l}
\toprule
\textbf{ID} & \multicolumn{1}{|c|}{\textbf{Distortion}} & \multicolumn{1}{|c|}{\textbf{Parameter}} & \textbf{Levels}  \\
\midrule
\multirow{2}{*}{1} & \multicolumn{1}{c|}{Temporal Shift} & \multicolumn{1}{c|}{shift length} &  \multirow{2}{*}{-1, -.5, -.125, .045, .1, .125, .25, .5, 1, 2} \\
& \multicolumn{1}{c|}{Audio} & \multicolumn{1}{c|}{(sec)} &   \\
\rowcolor{gray!20}
\multirow{2}{*}{2} & \multicolumn{1}{c|}{Audio Speed} & \multicolumn{1}{c|}{speed factor} & \multirow{2}{*}{.025, .05, .10, .15, .20, .25, .30, .4, .5, .75} \\
\rowcolor{gray!20} \multirow{-2}{*}{2} & \multicolumn{1}{c|}{Change up} & \multicolumn{1}{c|}{\%}  & \multirow{-2}{*}{.025, .05, .10, .15, .20, .25, .30, .4, .5, .75} \\
\multirow{2}{*}{3} & \multicolumn{1}{c|}{Video Speed} & \multicolumn{1}{c|}{speed factor} & \multirow{2}{*}{.025, .05, .10, .15, .20, .25, .30, .4, .5, .75} \\
& \multicolumn{1}{c|}{Change up} & \multicolumn{1}{c|}{\%} & \\
\rowcolor{gray!20}
\multirow{2}{*}{4} & \multicolumn{1}{c|}{Audio Speed} & \multicolumn{1}{c|}{speed factor} & \multirow{2}{*}{.025, .05, .10, .15, .20, .25, .30, .4, .5, .75} \\
\rowcolor{gray!20} \multirow{-2}{*}{4} & \multicolumn{1}{c|}{Change down} & \multicolumn{1}{c|}{\%} & \multirow{-2}{*}{.025, .05, .10, .15, .20, .25, .30, .4, .5, .75}  \\
\multirow{2}{*}{5} & \multicolumn{1}{c|}{Video Speed} & \multicolumn{1}{c|}{speed factor} & \multirow{2}{*}{.025, .05, .10, .15, .20, .25, .30, .4, .5, .75} \\
& \multicolumn{1}{c|}{Change down} & \multicolumn{1}{c|}{\%} &  \\
\rowcolor{gray!20}
\multirow{2}{*}{6} & \multicolumn{1}{c|}{Intermittent} & \multicolumn{1}{c|}{duration (sec) of mute}  & \multirow{2}{*}{.01, .025, .05, .1, .2, .3, .5, 1, 2.5, 4} \\
\rowcolor{gray!20} \multirow{-2}{*}{6} & \multicolumn{1}{c|}{Muting} & \multicolumn{1}{c|}{for every 1 sec} & \multirow{-2}{*}{.01, .025, .05, .1, .2, .3, .5, 1, 2.5, 4} \\
\multirow{2}{*}{7} & \multicolumn{1}{c|}{Randomly Sized} & \multicolumn{1}{c|}{gap duration (sec)}& \multirow{2}{*}{.01, .025, .05, .1, .2, .3, .5, 1, 2.5, 4} \\
 & \multicolumn{1}{c|}{Gaps Video} & \multicolumn{1}{c|}{and prob. 40\%} &  \\
\rowcolor{gray!20}
\multirow{2}{*}{8} & \multicolumn{1}{c|}{Fragment} & \multicolumn{1}{c|}{duration (sec) of}& \multirow{2}{*}{.3, .4, .5, 1, 1.5, 2, 2.5, 3, 3.5, 4} \\
\rowcolor{gray!20} \multirow{-2}{*}{8} & \multicolumn{1}{c|}{Shuffling} & \multicolumn{1}{c|}{segments to shuffle} & \multirow{-2}{*}{.3, .4, .5, 1, 1.5, 2, 2.5, 3, 3.5, 4} \\
\multirow{2}{*}{9} & \multirow{2}{*}{AV Flickering} & \multicolumn{1}{c|}{gap duration (sec)}& \multirow{2}{*}{.01, .025, .05, .1, .2, .3, .5, 1, 2.5, 4} \\
 &  & \multicolumn{1}{c|}{and prob. 40\%} &  \\
\bottomrule
\end{tabularx}
\caption{Summary of Distortions and their Parameters for Audio-Visual Content.}
\label{tab:type_values}
\end{table*}

\section{AVS Benchmark Dataset}\label{section:dataset}

To develop an effective metric that aligns closely with human perception, it is important to have a dataset representative of human judgments. With this in mind, our study was designed to create a benchmark for our metric training and to assess its correlation with human perception, especially in the context of AV synchronization.





\subsection{Data Preparation and Distortions} 
\label{sec:avdistortions}

Audio-visual synchrony, the alignment of audio and visual elements, is important for an immersive multimedia experience. Even minor disruptions can notably degrade the user's experience, as witnessed in scenarios like watching a video where a mere mismatch in drums playing (visuals) and beats (audio) can become distracting. Various real-world challenges, from unreliable network connection to encoding mishaps, can lead to such anomalies.
To produce realistic data representing such scenarios, we first collected 200 videos from the AudioSet corpus' \emph{evaluation} split \cite{audioset} ensuring that there is an aspect of synchrony in each of the samples. The samples selected are examples of ``in the wild'' videos, and does not include any samples with faces. We only include videos from actions such as: car driving by, dogs barking, instruments being played, manual labor being done, etc.\footnote{Appendix \ref{sec:data_labels} contains a brief overview of top-30 label frequencies contained within the videos subset selected for our dataset.} Each video then underwent nine synchrony related distortions at ten varying levels as shown in Table ~\ref{tab:type_values}, totaling in 18,200 distorted videos. 

\noindent Following~\cite{unterthiner2019accurate}, we apply static and temporal noises in the AV content. 
\textbf{Temporal noise} is where the distortions in either or both modalities cause de-synchronization in the AV content:
\begin{itemize}
\item \textbf{Temporal Misalignment:} The audio track was offset, either forwards or backwards relative to the video \cite{ATSC2003}, creating de-synchronizations spanning from -1 to 2 seconds. 
\item \textbf{Audio Speed Change:} Similar to FAD~\cite{Kilgour_2019}, the audio playback speed was manipulated, without corresponding changes in the video, resulting in gradual de-synchronization in the AV content. Depending on the speed change (up/down), we either clip the audio or visual frames to keep the same duration as the original video. {In this setting, we ensure the sound characteristics is preserved by keeping essential features such as pitch and timbre while altering speed to minimize changes on sound texture.}
\item \textbf{Video Speed Change:} Similar to FVD~\cite{unterthiner2019accurate}, the speed of the video is adjusted while leaving the audio track untouched. This would cause the visuals to slowly move out of sync with the audio. As in the audio speed change, we clip the content to keep the same duration as the original one.
\item \textbf{Fragment Shuffling:} Similar to the global swap distortion in FVD~\cite{unterthiner2019accurate}, both audio and video tracks were segmented and rearranged, preserving segment-level synchrony but disrupting overall synchronization of the video. 
\end{itemize}

\noindent \textbf{Static noise} is where the distortions do not necessarily change the synchrony in the AV content but it can be perceived as synchronization errors by humans:

\begin{itemize}
\item \textbf{Intermittent Muting:} Intermittent periods of silences with varying duration is introduced in the audio track. This distortion disrupts the continuity, which can pose a challenge for some synchronization metrics. This distortion is similar to Pops as described in Kilgour et. al.~\cite{Kilgour_2019}.
\item \textbf{Randomly Sized Gaps:} We randomly pick a frame with a certain probability and add black out visuals with varying number of frames in the video track. This distortion disrupts the continuity of the video. This distortion is similar to black rectangle noise as introduced in FVD.
\item \textbf{AV Flickering:} Periods of silence and blackout were introduced in the audio-visual track. Although flickering by itself is not a synchronization issue, it can however be perceived as synchronization errors by humans and poses a challenging scenario for metrics~\cite{itu1999subjective}. 
\end{itemize}


\subsection{Data Annotation} \label{sec:human_annotation}

Our aim in this task is to \textbf{NOT} evaluate the standalone quality of either audio or video. Instead, our focus is on synchrony issues and how humans perceive them.
Since pairwise stimulus experiments have shown to be faster than single-stimuli~\cite{clark_why_2018} and have been used in dubbing evaluation to foster relative measures in the rating~\cite{agrawal-etal-2023-findings}, we also present our participants with two videos side-by-side. They are asked to play both videos, compare them, and rate each one on a scale from 1-5 based on the provided guidelines.\footnote{Annotation guidelines are shown in Appendix \ref{sec:annotation_guidelines}. Upon publication, we will release the data set and code for creating distorted samples.}\looseness-1

From 18.2K videos, we randomly sampled (with replacement) 20K video pairs for comparison while taking into account the different levels of distortion (c.f. Table~\ref{tab:type_values}).
Participants conducted pairwise comparisons on these videos and each pair was rated at least three times by different annotators. This resulted in a total of 60,000 pairwise ratings or 120,000 ratings for individual videos. Each video duration being at least 10 seconds, this amounts to 100+ hours of annotated videos with each video annotated at least 3 times.
When looking at the inter-annotator agreement, we divided our data based on the severity of disagreement between the annotators. If all annotators annotated different scores those samples were put in ``disagreement'' group which constitute nearly 10\% of the total annotated data. 
Within the disagreement group, we saw about 16\% of cases where disagreements were equal or higher than 3, e.g., one video was rated as 1/4/5 by three annotators. These high-disagreement samples were then sent back to the annotators for QA, after which we received their revised annotations. We are unable to directly calculate inter-annotator agreement (like Cohen’s or Fleiss Kappa), as we had a large pool of annotators (80+) and not all annotators annotated all samples. However, we do observe a Krippendorff’s alpha value of 0.71 for the ratings assigned to all the samples, which corresponds to a moderate agreement~\cite{krippendorff04}.\looseness-1 
 
After removing outliers and deduplication we are left with 15.2K videos out of 40K.\footnote{Since we randomly sampled with replacement, one video was annotated approx. 2.4 times by annotators, which led to duplication.} We divide this 15.2K into 10.6K for training, 2.3K for development and 2.3K for evaluation. While creating these splits, we ensured that no original files and their derived (i.e., distorted) files are present in more than one set.\looseness-1

\begin{figure}
    \centering
    \includegraphics[width=0.6\linewidth]{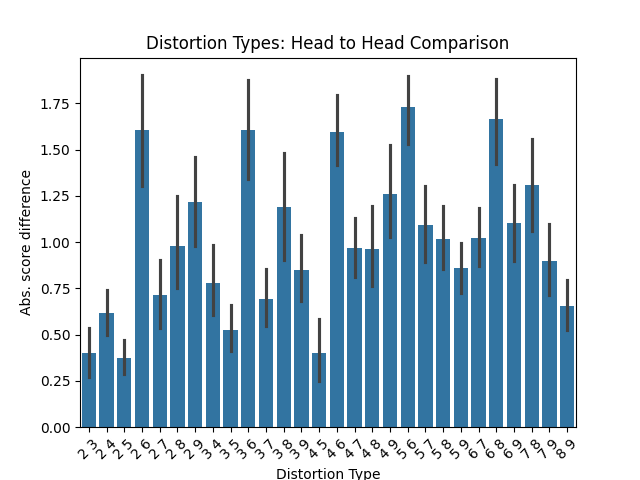}
    \caption{We compare absolute differences in annotation score across the distortion types. In this plot, x-axis shows distortion types that were compared, i.e., `2 4' represents distortion type 2 v/s 4 and y-axis represents the difference in scores across annotation tasks. For ID to distortion type mapping see Table~\ref{tab:type_values}.}
    \label{fig:data_analysis}
\end{figure}

To gain insights into which distortion types posed the most perceptual challenges for humans in terms of synchronization, we examined videos that underwent direct comparisons and analyzed the variations in the annotated scores.\footnote{We leave out audio-shift distortion in this analysis as the levels are not linearly increasing or decreasing.} Discrepancies in these scores (absolute values) offered insights into how differently these distortion types were perceived. Figure~\ref{fig:data_analysis} highlights that intermittent muting (type 6) is consistently the most noticeable distortion amongst all types (all high peaks). This observation aligns with the score distribution we observed in AVS benchmark (see Appendix Section~\ref{sec:avs_score_dist}) and our manual analysis where audio muting was the most disruptive distortion and humans could easily differentiate audio muting from other distortions.

\section{Preliminary Experiments}

Studies by Unterthiner \textit{et al.} \cite{unterthiner2019accurate} (Fréchet Video Distance) and Kilgour \textit{et al.} \cite{Kilgour_2019} (Fréchet Audio Distance) have demonstrated the efficacy of embeddings from I3D \cite{carreira2018quo} and VGGish \cite{hershey2017cnn} models in capturing qualitative and temporal characteristics from audio and video inputs, respectively. In the absence of dedicated metrics for assessing AV synchrony of `in the wild' content and the lack of baseline models, we formulated a new Fréchet audio-visual distance (FAVD) metric.\footnote{SparseSync~\cite{iashin2022sparse} and AVST~\cite{chen2021audiovisual} are only able to detect audio shift but our benchmark contain many more synchrony issues which render these metrics unusable.} Additionally, we want to assess if the combination of embeddings from I3D and VGGish models can be useful for detecting synchrony issues.

{Fréchet Distance metrics—FID, FAD, FVD, and now FAVD—compare the statistics of generated samples to those of the ground truth or a large collection of samples. In our preliminary experiment, we aimed to explore how FAD, FVD, and FAVD metrics respond to temporal discrepancies in the input data. We conducted our study on 18.2K videos, generating two sets of audio-visual embeddings: one from disrupted content and another representing aligned content (ground truth). For each type and level of distortion, we created a distinct "evaluation set" to analyze the impact of each specific distortion type and level on these metrics. 
In our experiments, we deliberately chose not to use FID (Fréchet Inception Distance) due to its limitation in evaluating temporal aspects. FID is designed for static images and cannot adequately assess content with time-dependent elements, such as videos, where scene continuity and flow are essential for quality evaluation.
By focusing on metrics capable of capturing these temporal dynamics, we aim to provide a more accurate and relevant evaluation of the audio-visual content's quality.}

We use I3D and VGGish to extract features from audio-visual content within the same time span (of 0.96 seconds - a default time span from VGGish). 

These representations are concatenated time-wise and we calculate multivariate Gaussians for both evaluation set embeddings, \(N_e(\mu_e, \Sigma_e)\), and the ground truth AV collection embeddings, \(N_{av}(\mu_{av}, \Sigma_{av})\). 
 
We then compute the Fréchet distance between these two Gaussians, as described by Dowson \textit{et al.} \cite{DOWSON1982450}: 

\begin{equation} \label{eq:frechet}
F(N_{av}, N_{e}) = \left| \mu_{av} - \mu_{e} \right|^2 + \text{\emph{tr}}(\Sigma_{av} + \Sigma_{e} - 2(\Sigma_{av} \Sigma_{e} )^{\frac{1}{2}})
\end{equation}
\noindent where \emph{tr} is the trace of the matrix. For FAD and FVD, we compute the embeddings from respective modalities.

\begin{figure}[ht]
    \centering

    \begin{minipage}{0.32\linewidth}
        \includegraphics[width=\linewidth]{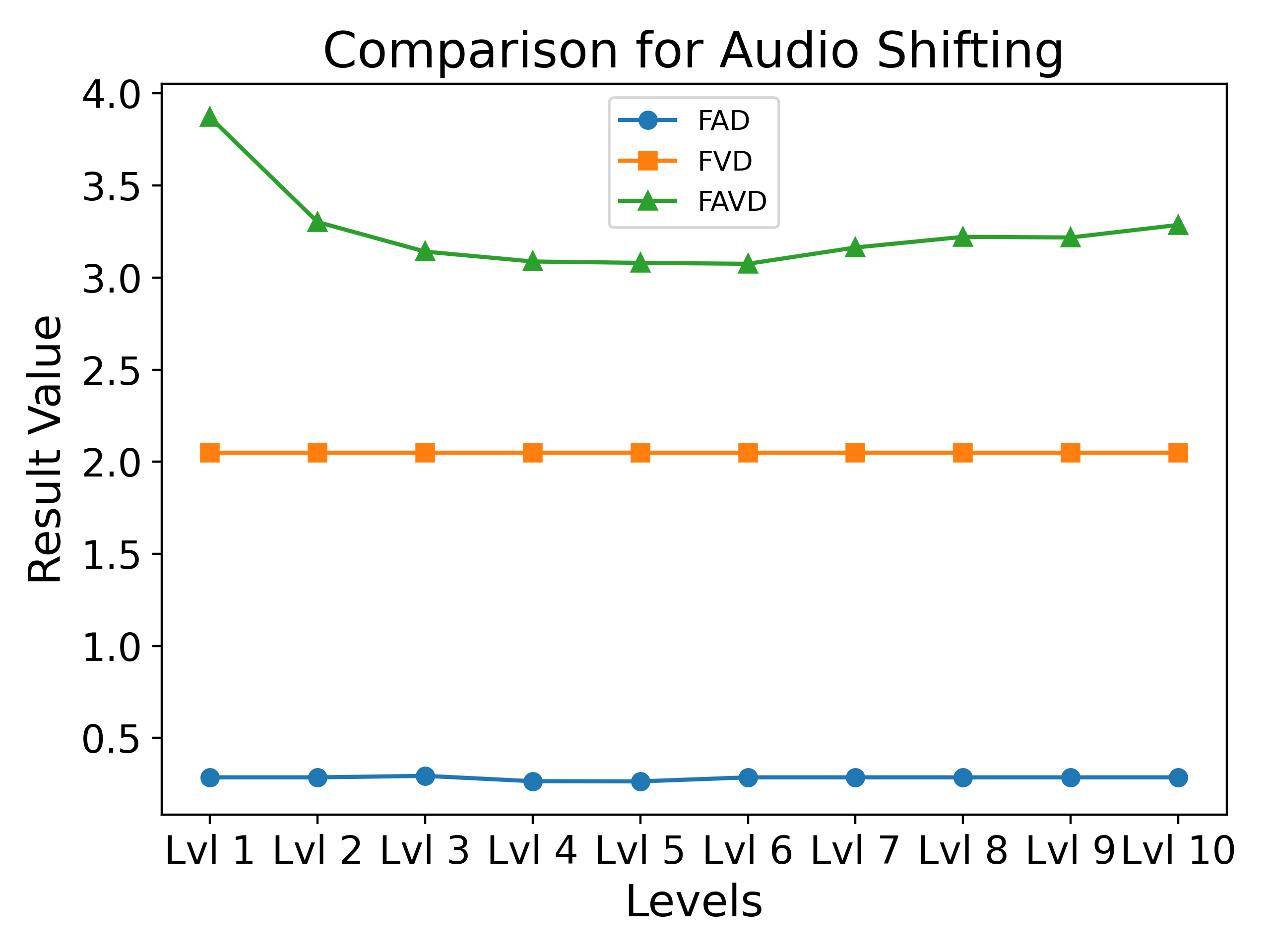}
        \subcaption{Audio Shift}
    \end{minipage}
    \hfill
    \begin{minipage}{0.32\linewidth}
        \includegraphics[width=\linewidth]{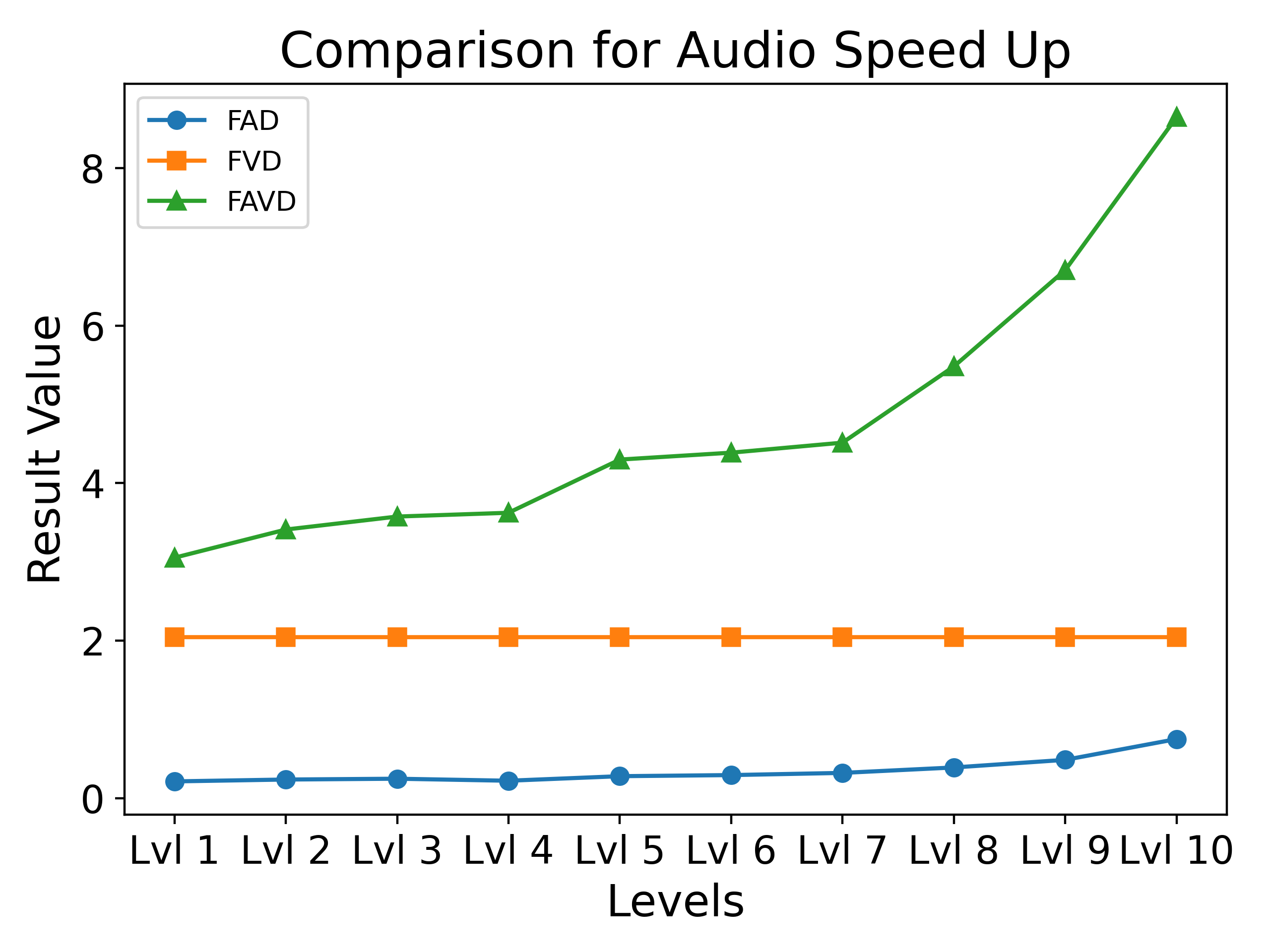}
        \subcaption{Audio Speed Up}
    \end{minipage}
    \hfill
    \begin{minipage}{0.32\linewidth}
        \includegraphics[width=\linewidth]{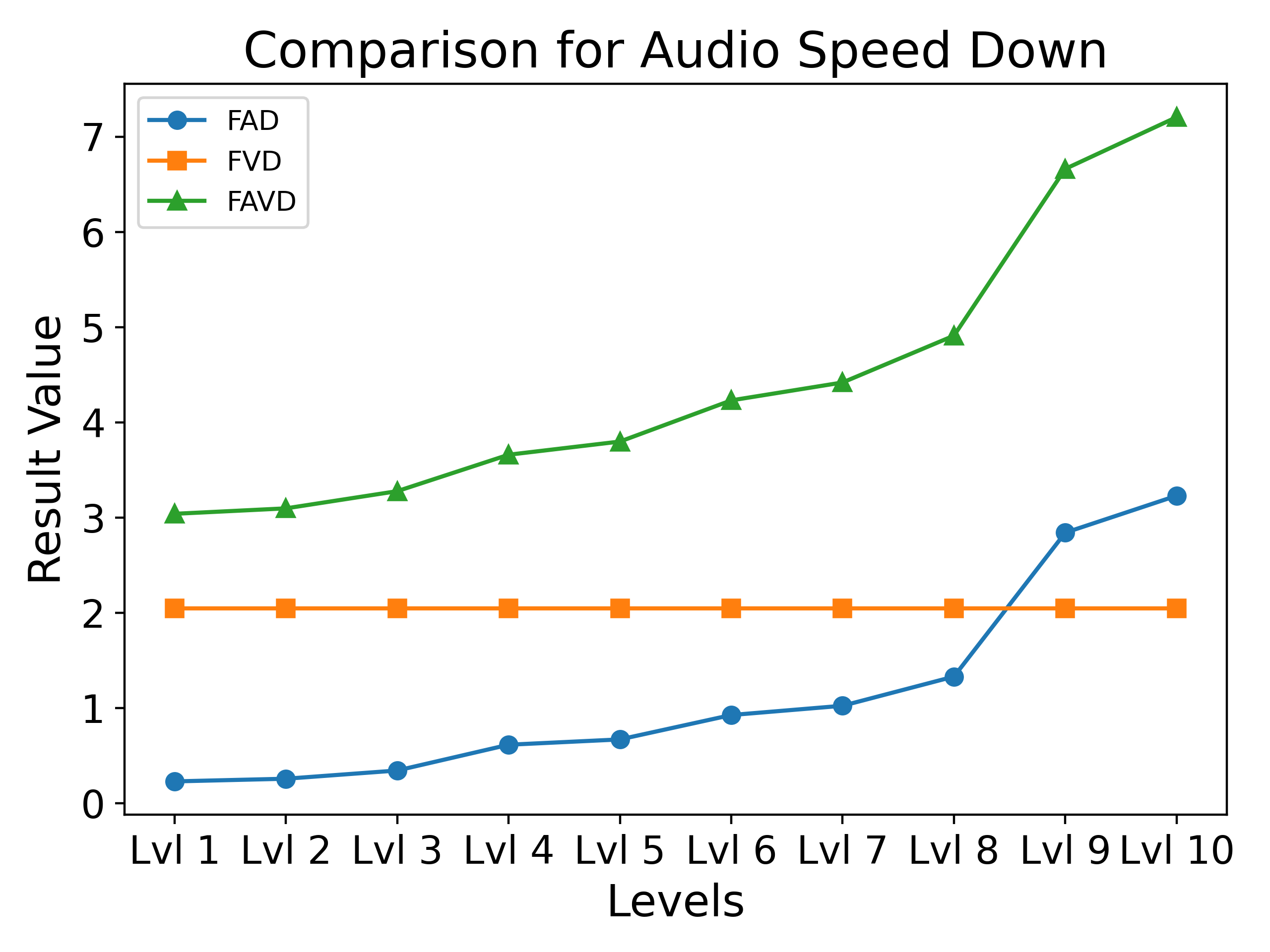}
        \subcaption{Audio Speed Down}
    \end{minipage}





    \caption{In these plots, we show the effect of distortions with varying levels on \textcolor{orange}{\textbf{FAD}}, \textcolor{blue}{\textbf{FVD}}, and \textcolor{green}{\textbf{FAVD}} for three distortion types. Flat trend line implies that a metric is not able to capture the distortion type. Increasing or decreasing trends show that a metric is susceptible to varying levels in a distortion. Distortion levels are taken from Table~\ref{tab:type_values}. Detailed plots for all distortion types are available in Appendix \ref{favd_interaction}.}
    \label{fig:fd_metrics}
\end{figure}

Examining the plots in Figure \ref{fig:fd_metrics}, it becomes evident that only FAVD (depicted in green) is responsive to various temporal distortions associated with AV synchrony across different levels. This is mainly due to use of covariance matrices ($\Sigma_{av}$, $\Sigma_{e}$) from audio-visual features (shown in Eq~\ref{eq:frechet}) which helps FAVD track the interaction of AV modality. Additionally, in cases like audio shift (a) where neither FAD nor FVD can detect synchronization issues (indicated by flat trend lines), and audio speed-up where FAD shows minimal movement, FAVD formulation exhibits a more pronounced reaction, highlighting that the fusion of features from I3D and VGGish models effectively captures temporality in an audio-visual context. Based on these observations, we chose to use I3D and VGGish features as inputs for our metric training, with FAVD serving as a baseline metric for the remainder of our work.

\section{PEAVS Score Formulation} \label{sec:peavs}
This section describes the architecture we have used to train our `PEAVS' score metric and its training details.

\subsection{Architecture}

\begin{figure*}[t]
    \centering
    \includegraphics[height=.24\textheight]{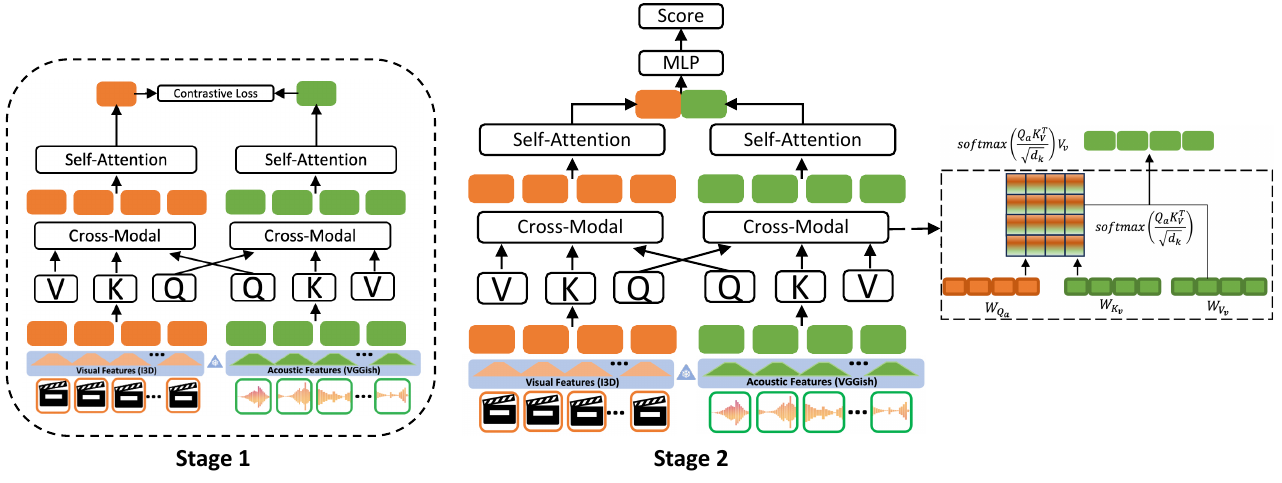}
    \caption{Framework Overview}
    \label{fig:Model}
\end{figure*}

The input to our proposed model are the video and audio streams, as seen in Figure \ref{fig:Model}. For feature extraction, we leverage I3D and VGGish.
In addition to this, our design draws inspiration from the audio-visual framework presented in \cite{Goncalves_2022}. Central to our architecture is the utilization of the cross-modal transformer formulation \cite{tsai2019MULT}, which excels in capturing temporal information and relationships between different modalities. To ensure temporal consistency across modalities, both audio and visual input features are extracted over matched time spans and are then relayed through cross-modal transformer layers.



There are several key components in the our architecture. First, the audio and visual inputs are processed through our frozen backbone feature extractors. Both features are extracted over 0.96 second windows. 
The I3D extracted inputs $x_v$ have dimensionality $x_v \in \mathbb{R}^{N_{av} \times  1024}$, where $N_{av}$ is the dimension of the audiovisual feature sequence, and 1,024 is the feature vector dimension. The acoustic feature vector $x_a$ extracted from the VGGish model is $x_a \in \mathbb{R}^{N_{av} \times 128}$, where $N_{av}$ is the dimension of the acoustic feature sequence, and 128 is the feature vector dimension. We ensure that all inputs to the cross-modal layers have same dimensionality, by projecting the visual features from 1,024 to 128 through the use of a 1D convolutional layer to produce $\bar{x}_v \in \mathbb{R}^{N_v\times128}$. 

Next, we introduce positional embeddings to the feature vectors, following the step adopted in the original transformer paper \cite{vaswaniattention}. 
The resultant features are used by the model to compute Query, Key, and Value vectors for each modality, as illustrated in Figure \ref{fig:Model}. In our model, the Query vector from one modality is paired in the transformer layer with the Key and Value vectors of the other, ensuring a cross-modal interaction. 
This design choice allows the cross-modal transformer encoder layers for each modality branch to produce representations for a given modality that are influenced by the other modality. Following this, the output of the cross-modal layers are refined through the self-attention layers to enhance the cross-modal representations. The resultant features are then processed by averaging the outputs from the self-attention layers followed by concatenation of these averaged representations from both modality branches. Finally, a \emph{multilayer perceptron} (MLP) with three fully-connected layers is integrated to produce a prediction score based on these concatenated features.

\subsection{Training}
As seen in Figure \ref{fig:Model}, our training method consists of a two-stage strategy. In stage 1, we pre-train the model using a contrastive learning task, distinguishing between aligned and non-aligned pairs. Afterwards, in Stage Two, we fine-tune the model on our primary task, which involves predicting scores based on human annotators scores.

\subsubsection{Stage 1}

In stage 1, we pre-train our model using a contrastive learning objective. In this phase, we used AudioSet's \emph{Balanced train} split \cite{audioset}, which after downloading all videos possible and removing corrupted ones resulted in 17K  videos for our contrastive training stage.\footnote{Note that this is different from AVS benchmark.} We then subjected each of these files to ten levels of temporal misalignment (audio shift), as specified in section \ref{sec:avdistortions}. This pre-processing resulted in a total of $\sim$190K files for the contrastive learning phase. Similar to SyncNet~\cite{Chung16a}, the training objective ensures that the outputs of both the audio and video branches are similar for aligned pairs and dissimilar for misaligned pairs. Specifically, we either minimize or maximize the Euclidean distance between the network outputs for these cases, respectively. Our objective consists of a contrastive loss (Equation \ref{eq:cont_loss}). The primary intention behind this pre-training is to embed regularization into the audio and visual features extracted from our frozen backbone networks.
This ensures that the model prioritizes temporal attributes of the features over their qualitative characteristics. 

\begin{equation}
\label{eq:cont_loss}
\mathcal{L}_c = \frac{1}{2N}\sum_{n=1}^{N}(y_n)D_n^2 + (1-y_n)(\max(0, m-D_n))^2
\end{equation}
where $y \in \{0,1\}$ is the label (1 for similar pairs and 0 for dissimilar pairs), \( D_w \) is the Euclidean distance between the two data points in the pair after passing them through the network,
and \( m \) is the margin.

\subsubsection{Stage 2}

In stage 2 the output modality heads of the pre-trained model are concatenated together to produce audio-visual representations. Three fully-connected layers are added on top of these concatenated heads. 
We then fine tune the entire model for the downstream task of predicting human-alignment scores. Since the aggregated human annotation scores are non-negative real numbers, we approach model training as a regression task and utilize \emph{concordance correlation coefficient} (CCC) to measure the agreement between the true and predicted scores. 
The CCC measurement is illustrated in Equation \ref{eq_ccc}.
\begin{equation}\label{eq_ccc}
\mathcal{L}_{CCC} =
\frac{2\rho\sigma_x\sigma_y}{\sigma_x^2+\sigma_y^2+(\mu_x-\mu_y)^2}
\end{equation}
\noindent where $\mu_x$
and $\mu_y$ denote the means of the true and predicted scores of a batch, respectively;
$\sigma_x$ and $\sigma_y$ represent the standard deviations of the true and
predicted scores of the batch, respectively; and $\rho$ is their Pearson's correlation coefficient.
Our training objective is to maximize CCC, ensuring that the predicted scores
closely correlate with the true scores.

\subsubsection{Training Details}

In stage 1, we divided the video files obtained from AudioSet's \emph{Balanced train} split \cite{audioset} into an 80\% training set and a 20\% validation set. This 80/20 split is carried out on the 17K original videos and extrapolated to distorted samples. We pre-trained the model in stage 1 until the validation loss plateaued. In stage 2, we use the train/validation/testing sets provided in the AVS benchmark corpus. We fine-tuned the model on training data for 20 epochs, saving the best-performing model based on the validation set performance.\footnote{For further details on the pre-training and fine-tuning settings, refer to appendix \ref{section:trainsetting}.}

\section{Experiments}

In this section, we evaluate the performance of our trained model using human-evaluation scores. We compare the model's scores with those provided by human evaluators on the \textbf{test split} of AVS benchmark. To further assess the effectiveness of our cross-modal transformer architecture for metric training, we run two ablation experiments.
First, to assess the impact of pre-training in stage 1, we directly train our model from scratch as in stage 2. This variant is termed ``Cross-Modal Base'' as it incorporates the same audiovisual cross-modal framework but excludes the contrastive learning pre-training phase.
Secondly, to assess the importance of cross-modal framework, we test a version without cross-modal layers called ``Basic Transformer''. This version employs only the self-attention branches, differently from the cross-modal approach of sharing query vector from one modality to another.

\begin{figure*}[h]
    \centering
    \begin{subfigure}[b]{\textwidth}
        \centering
        \includegraphics[height=.15\textheight]{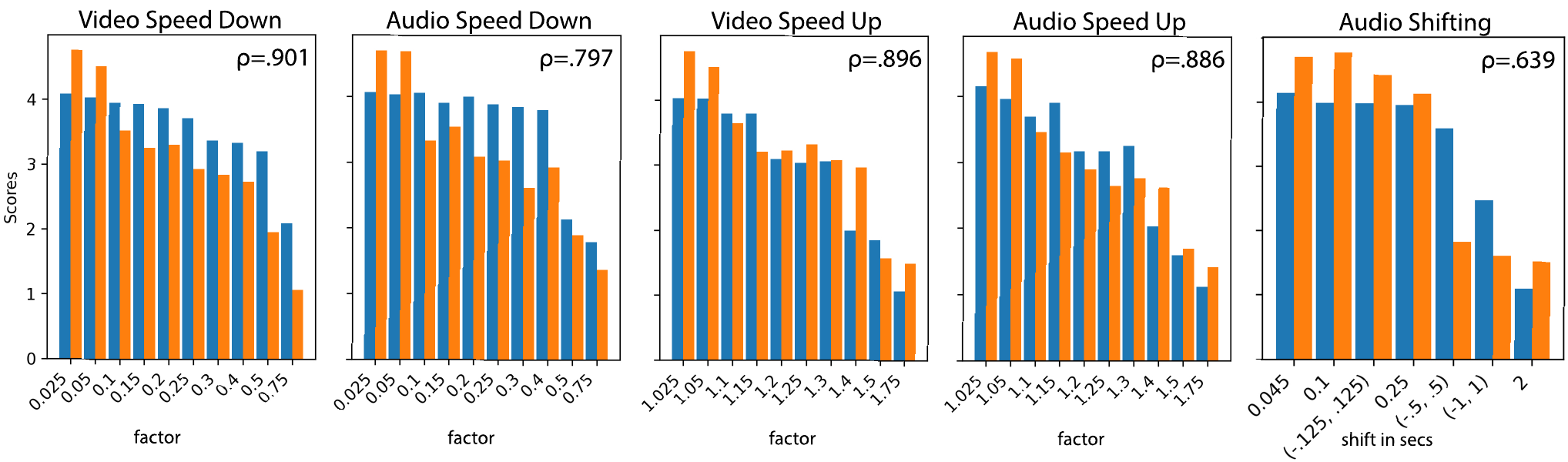}
    \end{subfigure}
    \begin{subfigure}[b]{\textwidth}
        \centering
        \includegraphics[height=.15\textheight]{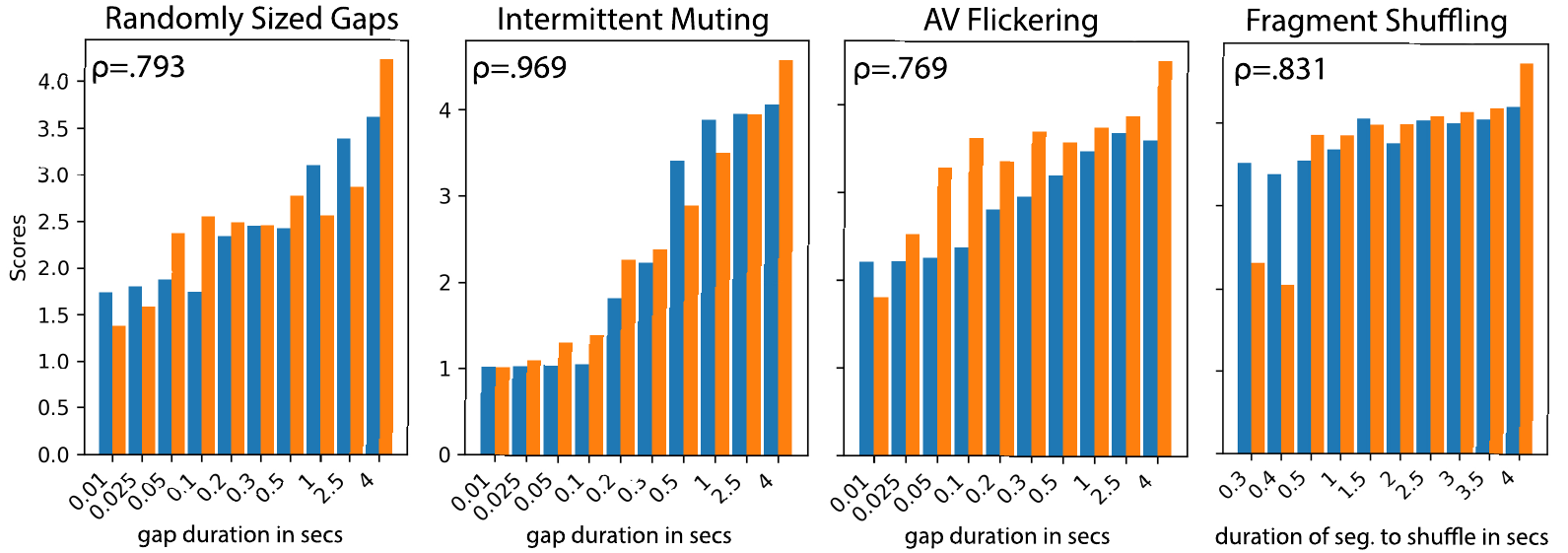}
    \end{subfigure}
    \caption{PEAVS results with Human Evaluation Scores compared side-by-side for each distortion type and at different levels. PEAVS scores are represented by \textcolor{blue}{\textbf{blue}} bars and Human Evaluation scores are represented by \textcolor{orange}{\textbf{orange}} bars. The x-axis are marked by the parameter and its levels as shown in Table \ref{tab:type_values}. $\rho$ is the correlation of PEAVS with human judgements.}
    \label{fig:metrics_distortions}
\end{figure*}

\subsection{Human Scores Correlation}
For the above models, we compute the correlation of their outputs with the human judgements using pearson correlation coefficient (PCC). Since Fréchet based metrics require a set of videos to calculate distance between multivariate Gaussians we compute set level correlations. As PEAVS and its variants produce a score for each video separately, we average the scores across the test set to compute set level scores.\footnote{Each set is defined as a collection of videos showing same distortion type and level. In total we had 90 sets, i.e. 9 distortions x 10 levels.} We also provide correlations at a much fine-grained clip level. Examining the results presented in Table \ref{tab:model_scores}, FAVD shows a relative improvement of 11\% (from 0.475 to 0.527) and 64\% (from 0.321 to 0.527) over FAD and FVD baselines setting a strong baseline for future work. This highlights FAVD's effectiveness in capturing AV synchrony issues.
PEAVS model achieves a Pearson correlation of 0.794 at the Set Level, marking a substantial 51\% improvement over the robust FAVD baseline. Notably, when we analyze the impact of pre-training on PEAVS (c.f. Cross-Modal Base), performance drops by 7\% \& 11\% at set and clip level, respectively, underscoring the significance of pre-training in PEAVS. Furthermore, removing the cross-modal layers (c.f. Base Transformer) results in a performance decrease of 9.4\% \& 19\% at set and clip level, respectively, emphasizing the critical role of cross-modality layers.


\begin{table}[h]
    \centering
    \begin{tabular}{l|c|c}
        \hline
        \textbf{Model Type} & \textbf{Set Level} & \textbf{Clip Level} \\ \hline
        FVD~\cite{unterthiner2019accurate} & 0.321 & - \\ \hline
        FAD~\cite{Kilgour_2019} & 0.475 & - \\  \hline
        FAVD (Ours) & 0.527 & - \\ \hline
        Basic Transformer & 0.720 & 0.434 \\ \hline
        Cross-Modal Base & 0.736 & 0.477 \\ \hline
        PEAVS (Ours) & \textbf{0.794} & \textbf{0.536} \\
        \hline
    \end{tabular}
    \caption{Correlation scores of different model types at Set and Clip levels.}
    \label{tab:model_scores}
\end{table}



\subsection{Per Distortion Analysis}

In this section, we analyze the response of PEAVS to varying distortions at different levels. Figure \ref{fig:metrics_distortions} factorizes the results in Table~\ref{tab:model_scores}, by each distortion type and level. PEAVS scores are shown side-by-side with the scores from human annotations on AVS benchmark test set. Generally, PEAVS metric aligns well with scores from human judgements across the board except in a few cases.

In top-right corner of Figure \ref{fig:metrics_distortions}, we present the effects of audio shift. For ease of analysis, we display these results in terms of absolute shift values. We start with a shift of 0.045 based on a report from ITU which indicated that the threshold for audio-visual shift detection by human ranges from -125ms to +45ms.\cite{ATSC2003}. At this level, both the metric's output and human scores peak ($>$4), and they subsequently decrease as the shift increases. 

In contrast to audio-shift, PEAVS shows the highest correlation for intermittent muting distortion which implies it is easier for the metric to capture this distortion. This confirms the notion that we observed in Section~\ref{sec:human_annotation} where intermittent muting stood out as the most noticeable distortion and it was easier for humans to detect it as well.

In case of the fragment shuffling (bottom-right corner), initial levels of perturbation --- shuffled fragments of 0.3-0.4 seconds --- is significantly more disruptive for humans than the metric, i.e. the metric fails to capture this insight. 



\begin{table}[h]
    \small
    \centering
    \begin{minipage}{.5\linewidth}
    \centering
    \begin{tabular}{cc|c|c}
    & & \multicolumn{2}{c}{Predicted} \\ \cline{3-4}
      &  & Pos. & Neg. \\ \hline
    \multirow{2}{*}{\rotatebox[origin=c]{90}{Actual}} &  \multicolumn{1}{|c|}{Pos} & 39 & 227 \\ \cline{2-4}
    & \multicolumn{1}{|c|}{Neg} & 4 & 130 \\
    \end{tabular}
    \end{minipage}%
    \begin{minipage}{.5\linewidth}
    \centering
    \begin{tabular}{cc|c|c}
    & & \multicolumn{2}{c}{Predicted} \\ \cline{3-4}
      &  & Pos. & Neg. \\ \hline
    \multirow{2}{*}{\rotatebox[origin=c]{90}{Actual}} & \multicolumn{1}{|c|}{Pos} & 92 & 174 \\ \cline{2-4}
    & \multicolumn{1}{|c|}{Neg} & 24 & 110 \\
    \end{tabular}
    \end{minipage}%
    \vspace{.2cm}
    \caption{Confusion matrix for SparseSync (left) and PEAVS (right). Accuracy of SparseSync = 42.3\% v/s PEAVS = 50.5\%}
    \label{tab:sparsesync_vs_peavs}
\end{table}

\vspace{-1cm}
\subsection{PEAVS vs. SparseSync}


In this section, we report an experiment comparing PEAVS with SparseSync, which was introduced by Iashin \textit{et al.} \cite{iashin2022sparse} for detecting audio-visual shifts inside `in the wild' videos. While SparseSync's goal is to objectively quantify specific synchronization issues, by estimating an audio-visual offset shift value, PEAVS evaluates synchronization under multiple facets and according to a perceptual scale. Due to their diverging objectives, a direct comparison is challenging. Moreover, SparseSync is trained on 21 classes representing distinct audio-shift levels from -2.0 sec (left) to 2.0 sec (right) with increments of 0.2, with 0.0 representing no audio shift; while PEAVS is trained to predict float values between 1 and 5. 

In this experiment, we assess both model's performance in accurately classifying 400 videos, including 200 ground truth videos and 200 randomly audio-shifted versions of them. \footnote{More details about this evaluation set are in Appendix Section~\ref{sec:sparsesyncvsavsync}.} 
We consider distortions of 0.045, 0.1 and $\pm$0.125 as positive cases (i.e. ground truth) as ITU recommend an acceptability threshold for AV shift up to 185 ms~\cite{noauthor_bt1359_nodate}. This also aligns well with SparseSync buckets where first distortion starts at $\pm$0.2 seconds (i.e. SparseSync's step-size). 
To ensure a fair assessment, we divided the PEAVS scale into 21 bins, matching the output structure of SparseSync.\footnote{We use increments of 0.238 (=$\frac{5}{21}$).}  For ground truth videos, we anticipate that SparseSync will classify them as having zero shifts (0.0 class), while PEAVS is expected to assign them to the highest scoring bin (i.e. $(4.76,5]$).

In Table~\ref{tab:sparsesync_vs_peavs}, we show the confusion matrices for SparseSync (left) and PEAVS (right). In terms of accuracy, PEAVS outperforms SparseSync by 19\%, i.e. 50.5\% vs.  42.3\%. 
In particular, SparseSync shows an 18\% edge in detecting distortions while PEAVS shows a 136\% higher rate in detecting clean content. 

\section{Limitations}
Like any research, our work is not without its limitations, and we have identified a couple of them here:

\noindent \textbf{Dataset}: Due to budget constraints (30K USD), we could only source 200 videos and conduct 60K pair-wise annotations. In selecting these 200 videos, we ensured diversity. Our dataset comprises 215 labels out of the 512 present in AudioSet, representing only 40\% of labels. Expanding this dataset for broader video coverage is left for future work.

\noindent \textbf{Metric}: The limited size of the AVS benchmark highlights the necessity of pre-training in model-based metrics (in our case, PEAVS). This is further evidenced by the results in Table~\ref{tab:model_scores}, where pre-training improves correlation from 0.736 (Cross-Modal Base) to 0.794 (PEAVS). In this study, we only pre-trained PEAVS on audio-shift noise, leaving pre-training on other noise types for future work.

\noindent \textbf{``in the wild'' videos}: Due to proprietary issues, we could not experiment with datasets featuring talking faces like VoxCeleb~\cite{Nagrani17} or Lip Reading Sentences~\cite{afouras2018lrs3ted}. Our study is confined to addressing synchrony issues for ``in the wild'' videos.


\section{Conclusions}


We introduced PEAVS, a novel metric designed for measuring synchrony in audio-visual content, an important advancement in the field of evaluation of audio-visual content. While the domain of audio-visual generative modeling is ever-evolving, assessing synchronization remains crucial for enhancing user experience. Although numerous metrics cater to either audio or visual content separately, a noticeable gap exists in evaluating their synchronization with a focus on viewers opinions. Our study addresses this gap with PEAVS, a metric based on a vast human-annotated dataset. The strong correlation between PEAVS scores and human evaluations confirms its efficacy in capturing perceptions surrounding audio-visual synchronization. As the scope of audio-visual content generation expands, tools like PEAVS are essential in maintaining a balanced interplay between sight and sound, guiding both creators and researchers.

Furthermore, we introduced the Audio-Visual Synchrony human perception benchmark, which provides a new dataset detailing human-perceptual scores on various audio-visual synchrony challenges for real-world videos. By isolating specific challenges in a controlled audio-visual environment, we aim to simplify the task for researchers to propose, assess, and analyze potential solutions while maintaining agreement with human perception. 

\clearpage  

%
%
\bibliographystyle{splncs04}
\bibliography{main}

\clearpage
\appendix

\section{Model Training Settings}
\label{section:trainsetting}

\begin{table}[h]
\centering
\caption{Training Settings}
\begin{tabular}{l|l|l}
\hline
\textbf{Parameter} & \textbf{Stage 1} & \textbf{Stage 2}  \\
\hline
LR  & 0.001    & 0.0001   \\
\rowcolor{gray!20}Optimizer     & Adam      & Adam  \\
Loss    & Contrastive   & CCC    \\
\rowcolor{gray!20}Batch Size    & 128        & 64    \\
Epochs    & 60       & 20    \\
\rowcolor{gray!20}LR Scheduler  & .1 (Patience 10) & .1 (Patience 3)   \\
Margin    & 1.0       & -    \\
\hline
\end{tabular}
\end{table}

\begin{table}[h]
\centering
\caption{Transformer Layers Settings}
\begin{tabular}{l|l}
\hline
\textbf{Parameter}   & \textbf{Value} \\
\hline
Heads                    & 8     \\
\rowcolor{gray!20}Layers                   & 3     \\
Embed Dim                & 128   \\
\rowcolor{gray!20}Attention Dropout        & 0.1   \\
Relu Dropout             & 0.1   \\
\rowcolor{gray!20}Embed. Dropout           & 0.25  \\
Residual Block Dropout   & 0.1   \\
\hline
\end{tabular}
\end{table}

\section{Human Annotation Guidelines}
\label{section:exp}

\subsection{Background}
In this annotation task, we want to evaluate the synchronization quality between audio and visual modalities in the video, and get an understanding of how the two videos presented to you compare to each other in terms of quality of synchronization. Please note that:

\begin{itemize}
    \item We are NOT looking at the perceived quality of audio or video independently, but only how well they are aligned.
    \item We added certain types of distortions in these videos that disrupt the AV synchrony like speeding up the audio or speeding up the video, intermittently muting the audio, blacking out parts of video, flickering, etc. 
    \item Remember that we are also interested to see how the synchronization quality in these videos compares to each other, i.e. your relative judgment.
\end{itemize}


\subsection{Annotation Criteria}
\label{sec:annotation_guidelines}

In each comparison view, you will see two videos side by side. Play both videos and rate them on a scale of 1-5  in terms of disruption caused by these distortions based on the following likert scale: 

\begin{enumerate}
    \item Score 1 when there is complete misalignment between audio and video OR the video/audio is totally incomprehensible due to disruptions.
    \item Score 2 when only a few parts of audio/video are in alignment OR large portion of video/audio is incomprehensible due to disruptions.
    \item Score 3 when there is moderate mis-alignment between audio/video OR some portion of video/audio is comprehensible but there are visible disruptions.
    \item Score 4 when there is almost perfect alignment with minor mis-alignments in some parts of video OR most of the video and audio is comprehensible with minor disruptions.
    \item Score 5 when there is perfect alignment and audio/video are flawlessly in sync, AND/OR have no disruptions at all.
\end{enumerate}

\section{Dataset Labels Information}
\label{sec:data_labels}

\begin{figure}
    \centering
    \includegraphics[width=0.7\linewidth]{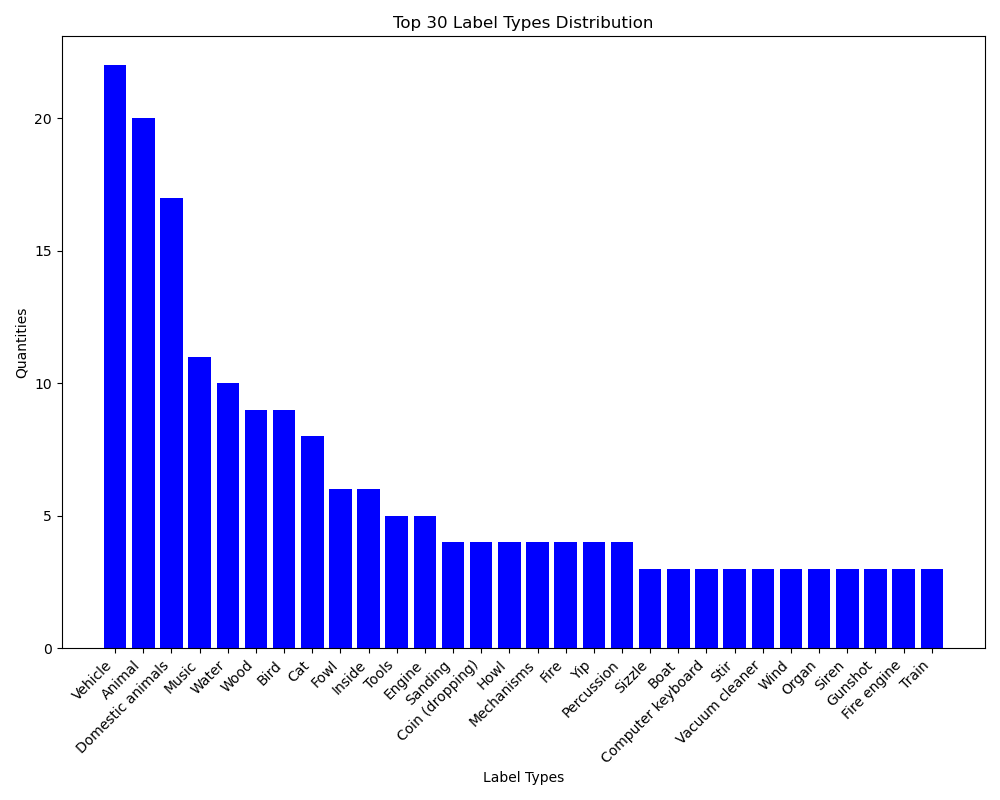}
    \caption{Histogram of the top 30 label frequencies from the AVS benchmark sub-sampled from the AudioSet corpus.}
    \label{fig:histogram_top30}
\end{figure}

To generate our dataset, we selected 200 videos from AudioSet \cite{audioset} ensuring that each sample contained well-aligned audio-visual content. The chosen samples are representative of videos "in the wild"; we excluded any samples that featured talking faces. Instead, our selection primarily contains videos depicting actions such as: cars driving by, dogs barking, instruments being played, and manual labor, among others. Based on the labels provided in AudioSet, our dataset comprises 215 labels. The number of labels exceeds the number of files because some files may be associated with multiple classes, resulting in co-occurrence of labels in many files within AudioSet. To provide an overview of the labels in our selected dataset, Figure \ref{fig:histogram_top30} presents a histogram of the top 30 label frequencies from our AudioSet subset.

\begin{figure}[t!]
    \centering
    \begin{subfigure}[t]{0.5\textwidth}
    \centering
        \includegraphics[height=0.7\textwidth]{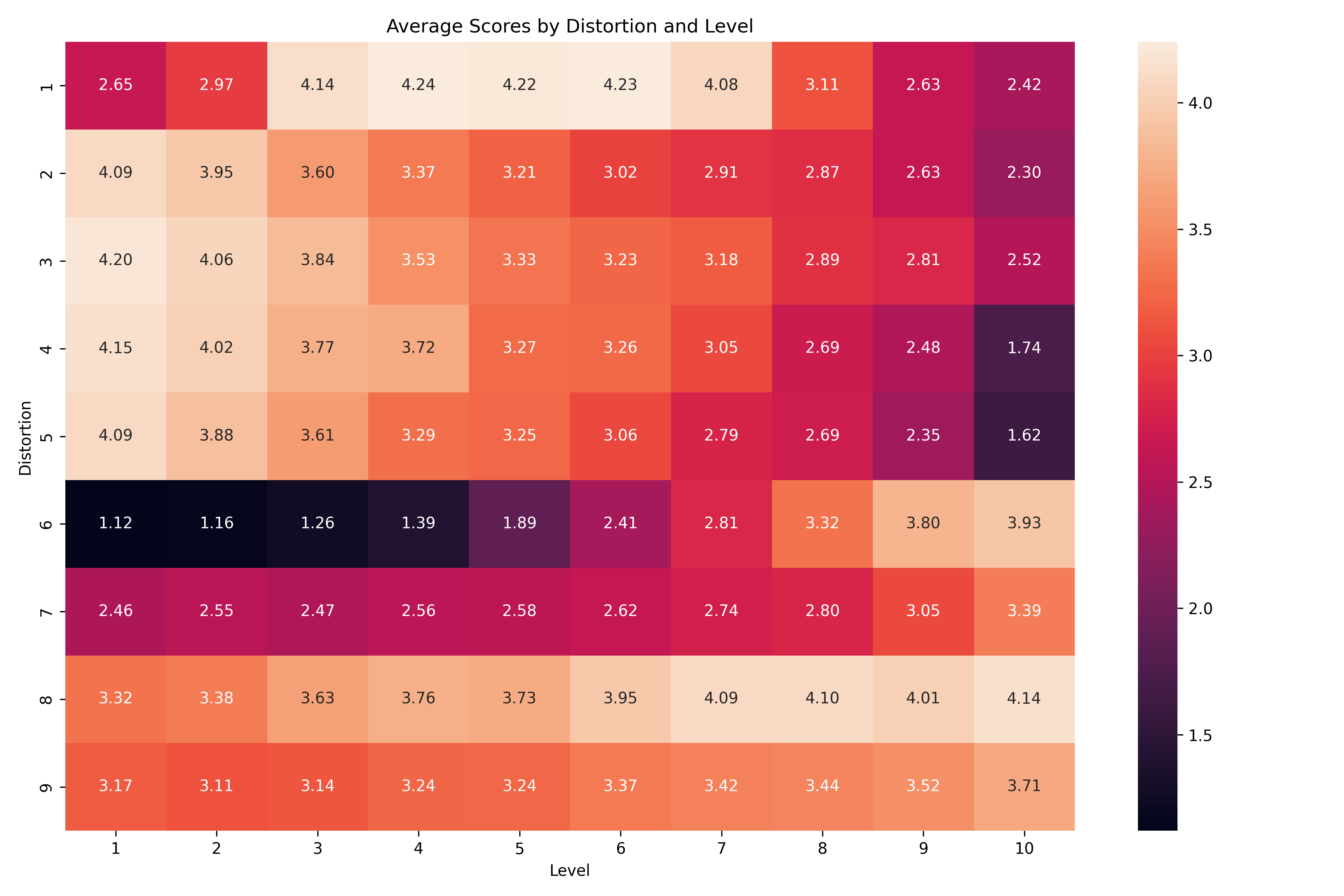}
    \end{subfigure}\hfil
    \begin{subfigure}[t]{0.5\textwidth}
    \centering
        \includegraphics[height=0.7\textwidth]{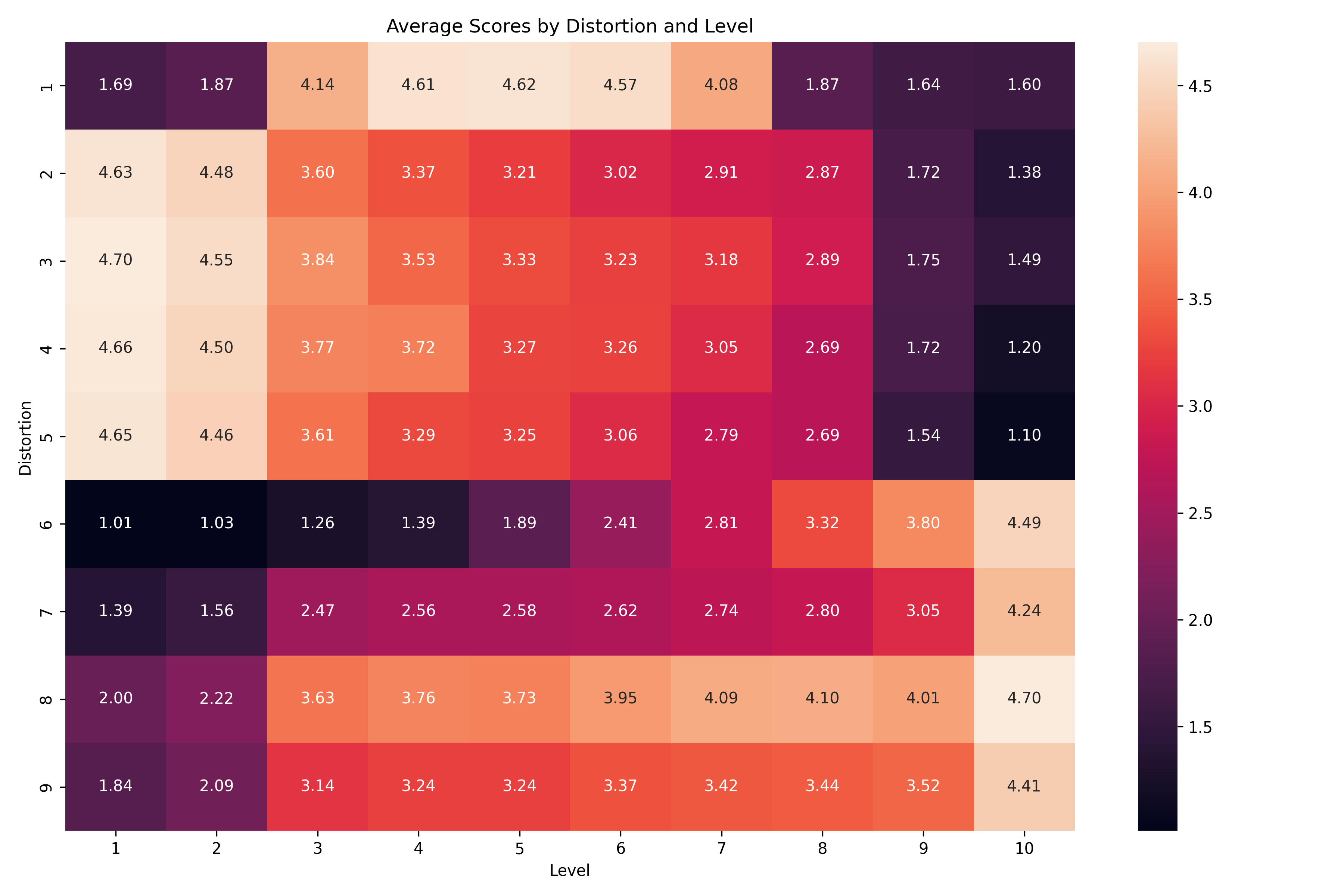}
    \end{subfigure}
    \caption{Heatmap of score distribution on AVS benchmark, across distortion types and levels before (left) and after (right) filtering of data. The lighter the color the higher the scores and vice-versa.}
    \label{fig:heatmap}
\end{figure}

\section{Score Distribution in AVS Benchmark}
\label{sec:avs_score_dist}
In this section, we will analyse the score distribution on the AVS benchmark data and the effect of filtering. 
We wanted to do a followup analysis from the main paper, especially by looking at the distribution of human annotation scores in the AVS benchmark data. Figure~\ref{fig:heatmap} provides more insights in our benchmark. Here, we observe a similar trend of intermittent muting (distortion type 6) being highly disruptive perceptually (colored with dark cells in the figure) compared to other distortion types especially in the lower levels of distortion - where the distorted videos are more jittery. 

To remove the outliers in the benchmark, we filtered the dataset on the following criteria:
\begin{enumerate}
    \item We filter out all samples where ground truth videos are annotated with an average score (of 3 raters) of 3.5 or below.
    \item We filter out all samples with extreme distortion levels rated as 5. 
\end{enumerate}
Filtering the data increases the variance within each distortion type and a clear trend across levels emerges. After filtering, less distorted videos have higher scores as compared to those without filtering. See distortion type IDs 7,8,9 (in Figure~\ref{fig:heatmap}) where values in initial level shows an increasing trend in the right plot (filtered) as opposed to the left one (unfiltered). 

Note that in Figure~\ref{fig:heatmap}, distortion type IDs are the same as in Table~\ref{tab:type_values}. Levels in the audio-shift distortion is sorted in increasing order (from left shift of -1 to right shift of +2).

\section{FAD v/s FVD v/s FAVD} \label{favd_interaction}

\begin{figure}[ht]
    \centering

    \begin{minipage}{0.32\linewidth}
        \includegraphics[width=\linewidth]{images/frechet_dist_line_plots/Audio_Shifting.png}
        \subcaption{Audio Shift}
    \end{minipage}
    \hfill
    \begin{minipage}{0.32\linewidth}
        \includegraphics[width=\linewidth]{images/frechet_dist_line_plots/Audio_Speed_Up.png}
        \subcaption{Audio Speed Up}
    \end{minipage}
    \hfill
    \begin{minipage}{0.32\linewidth}
        \includegraphics[width=\linewidth]{images/frechet_dist_line_plots/Audio_Speed_Down.png}
        \subcaption{Audio Speed Down}
    \end{minipage}

    \vspace{0.2cm}

    \begin{minipage}{0.32\linewidth}
        \includegraphics[width=\linewidth]{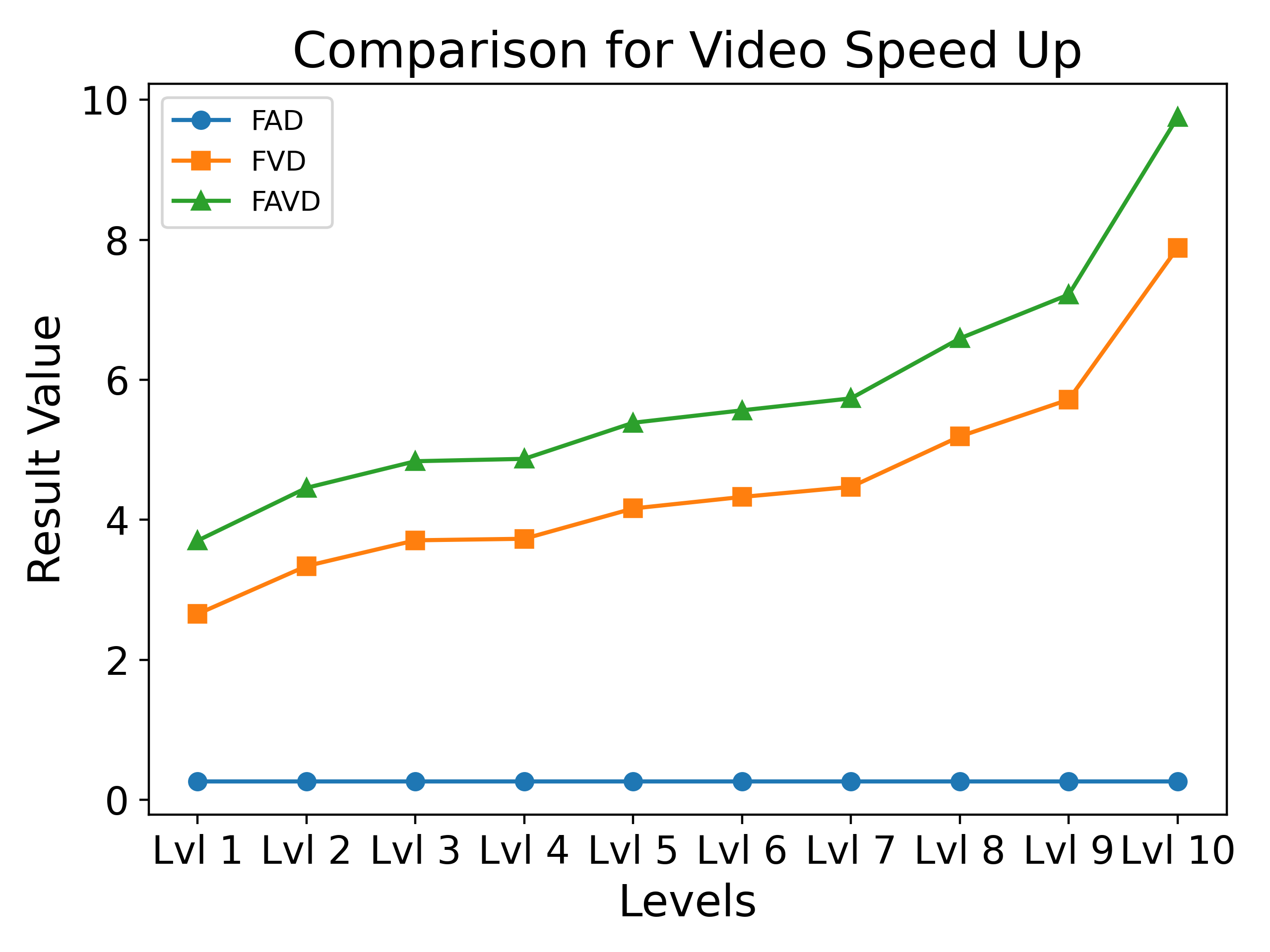}
        \subcaption{Video Speed Up}
    \end{minipage}
    \hfill
    \begin{minipage}{0.32\linewidth}
        \includegraphics[width=\linewidth]{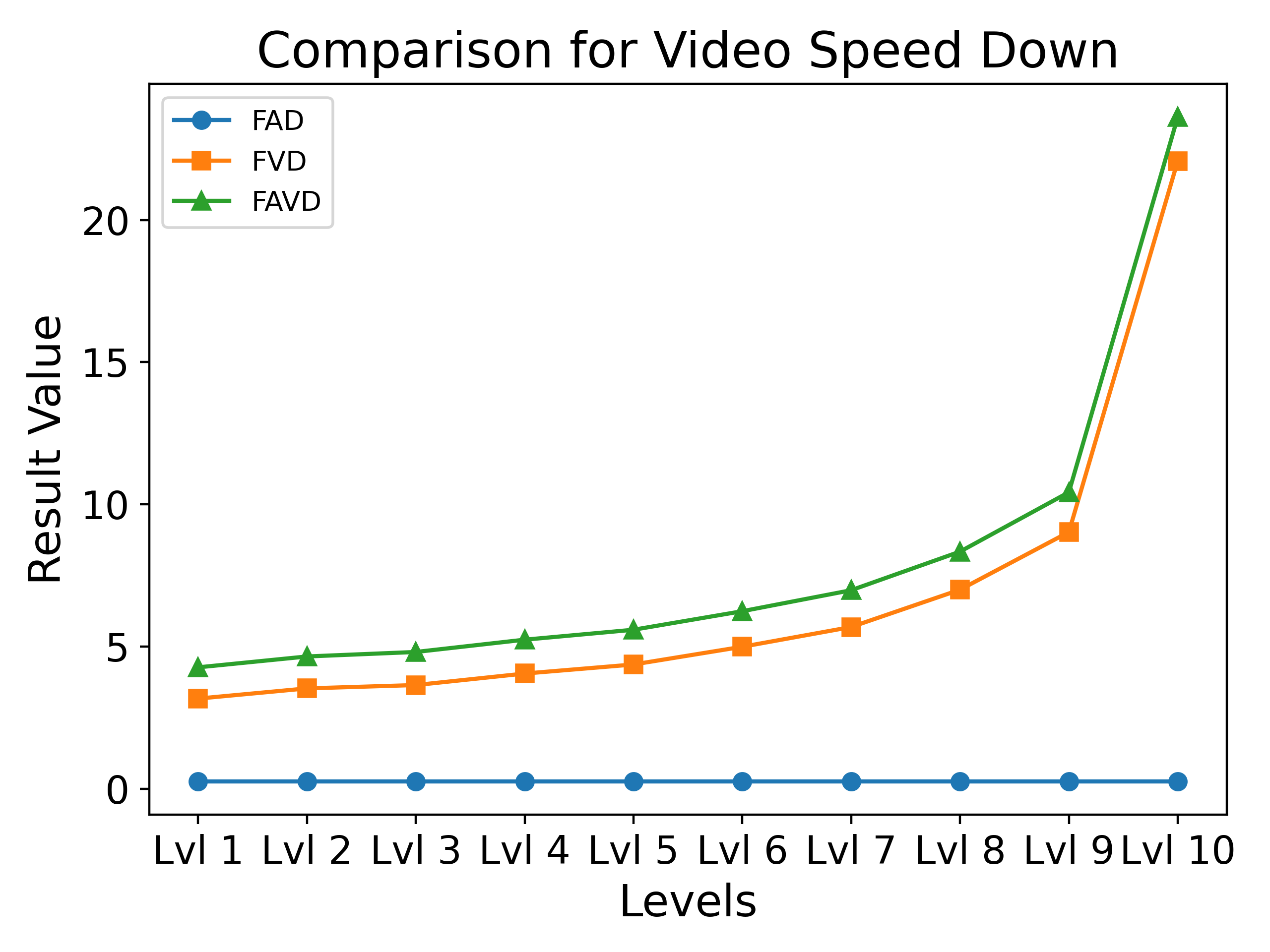}
        \subcaption{Video Speed Down}
    \end{minipage}
    \hfill
    \begin{minipage}{0.32\linewidth}
        \includegraphics[width=\linewidth]{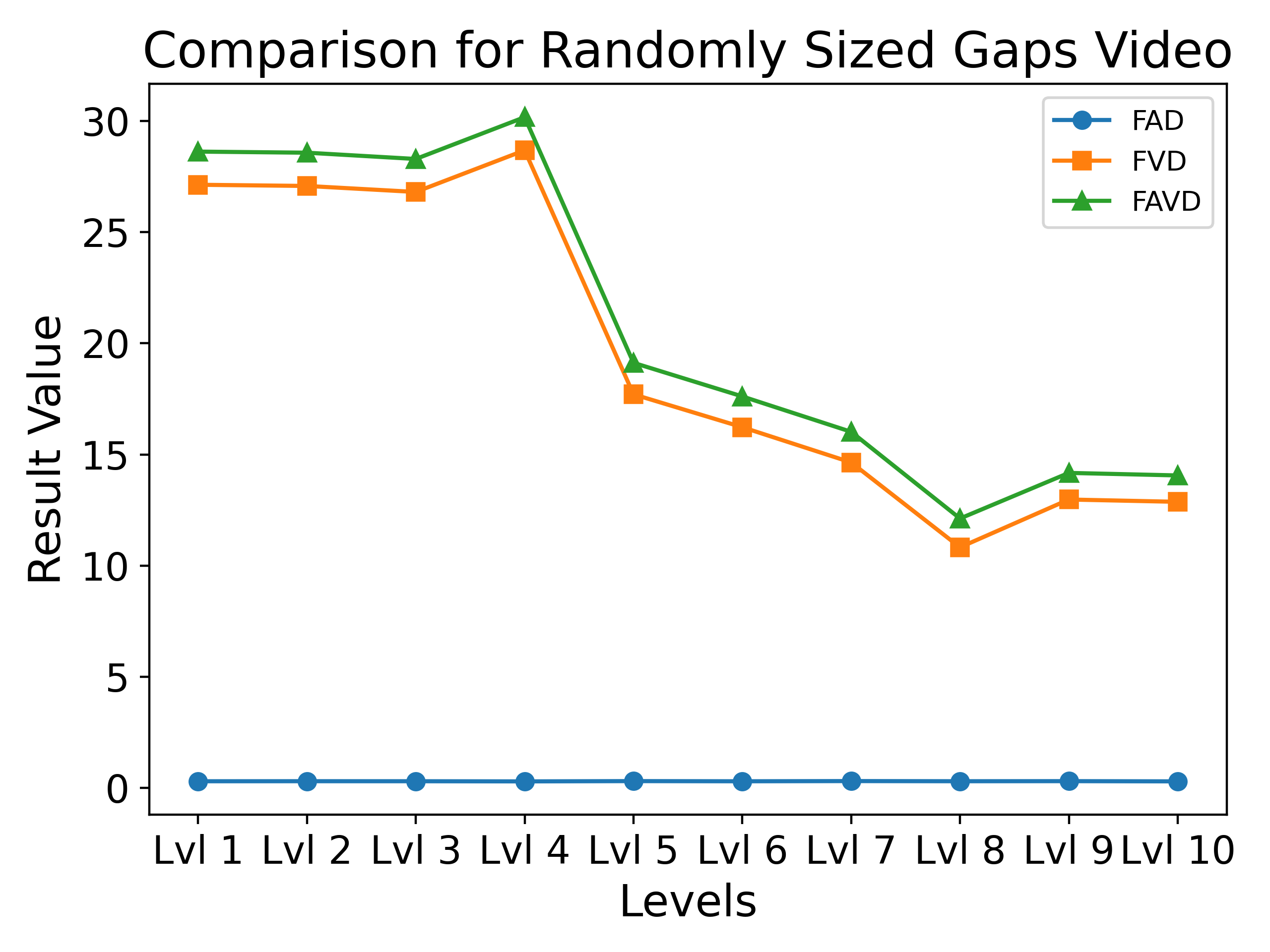}
        \subcaption{Random Gaps}
    \end{minipage}

    \vspace{0.2cm}

    \begin{minipage}{0.32\linewidth}
        \includegraphics[width=\linewidth]{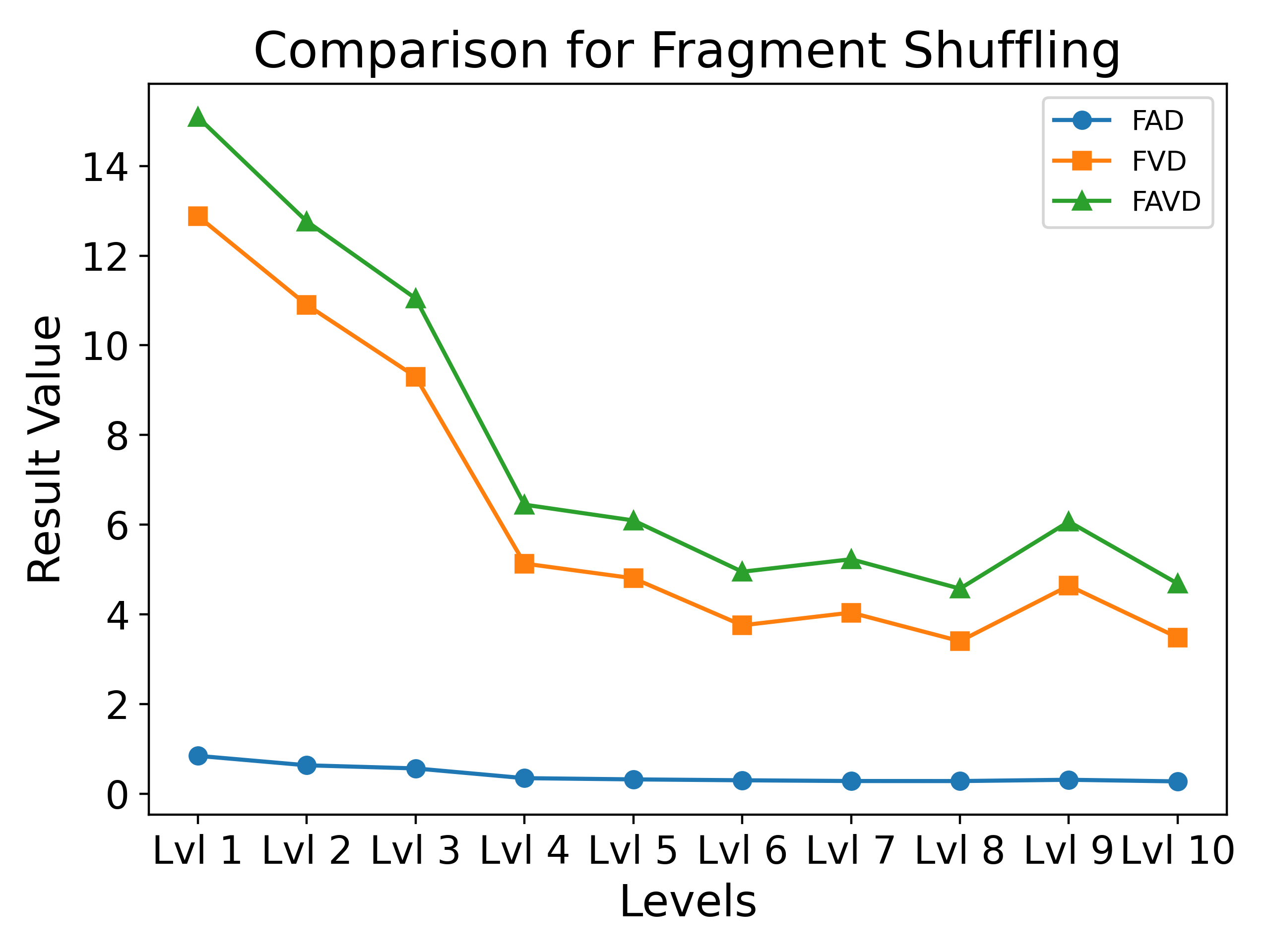}
        \subcaption{Fragment Shuffle}
    \end{minipage}
    \hfill
    \begin{minipage}{0.32\linewidth}
        \includegraphics[width=\linewidth]{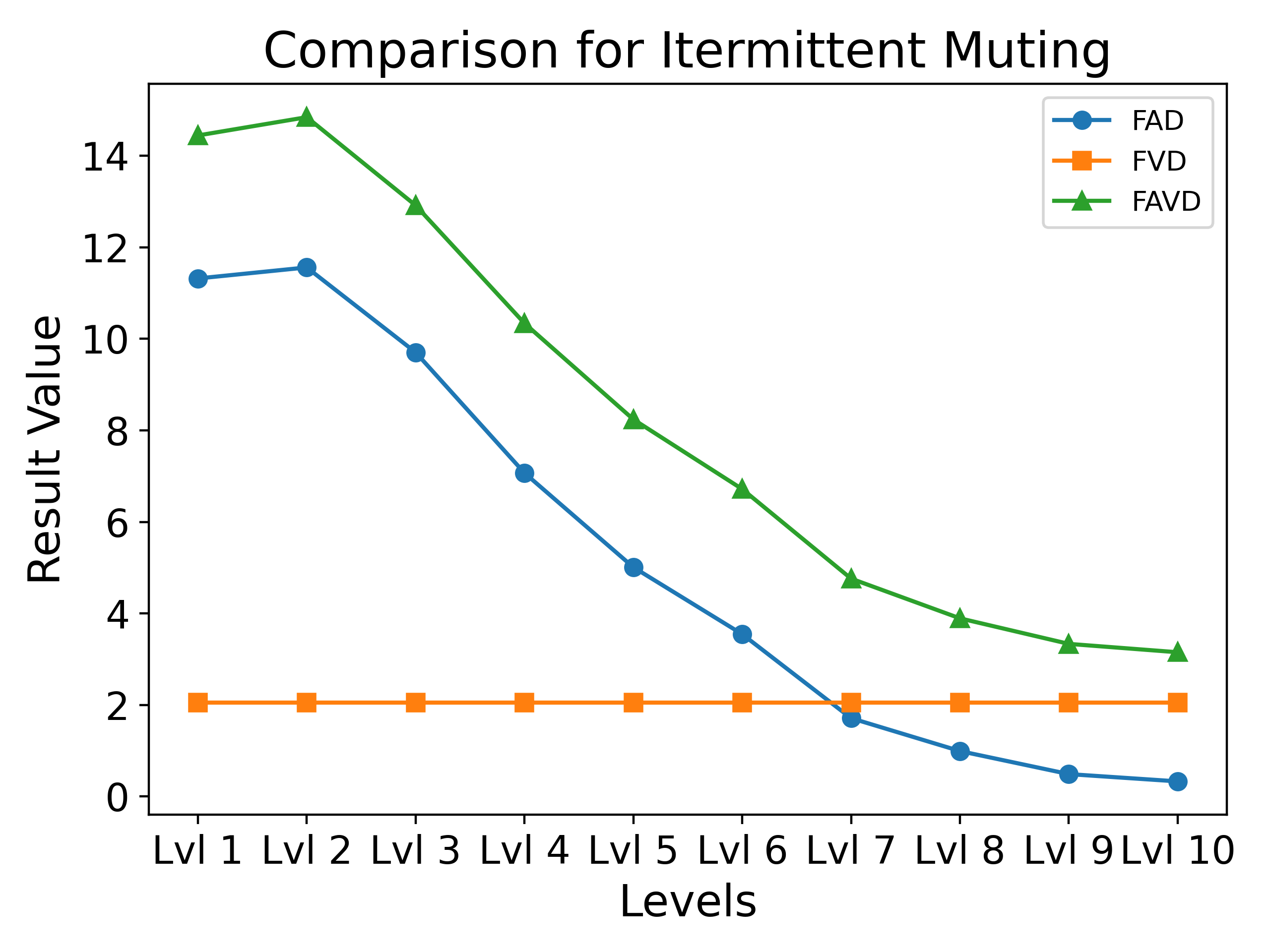}
        \subcaption{Intermittent Mute}
    \end{minipage}
    \hfill
    \begin{minipage}{0.32\linewidth}
        \includegraphics[width=\linewidth]{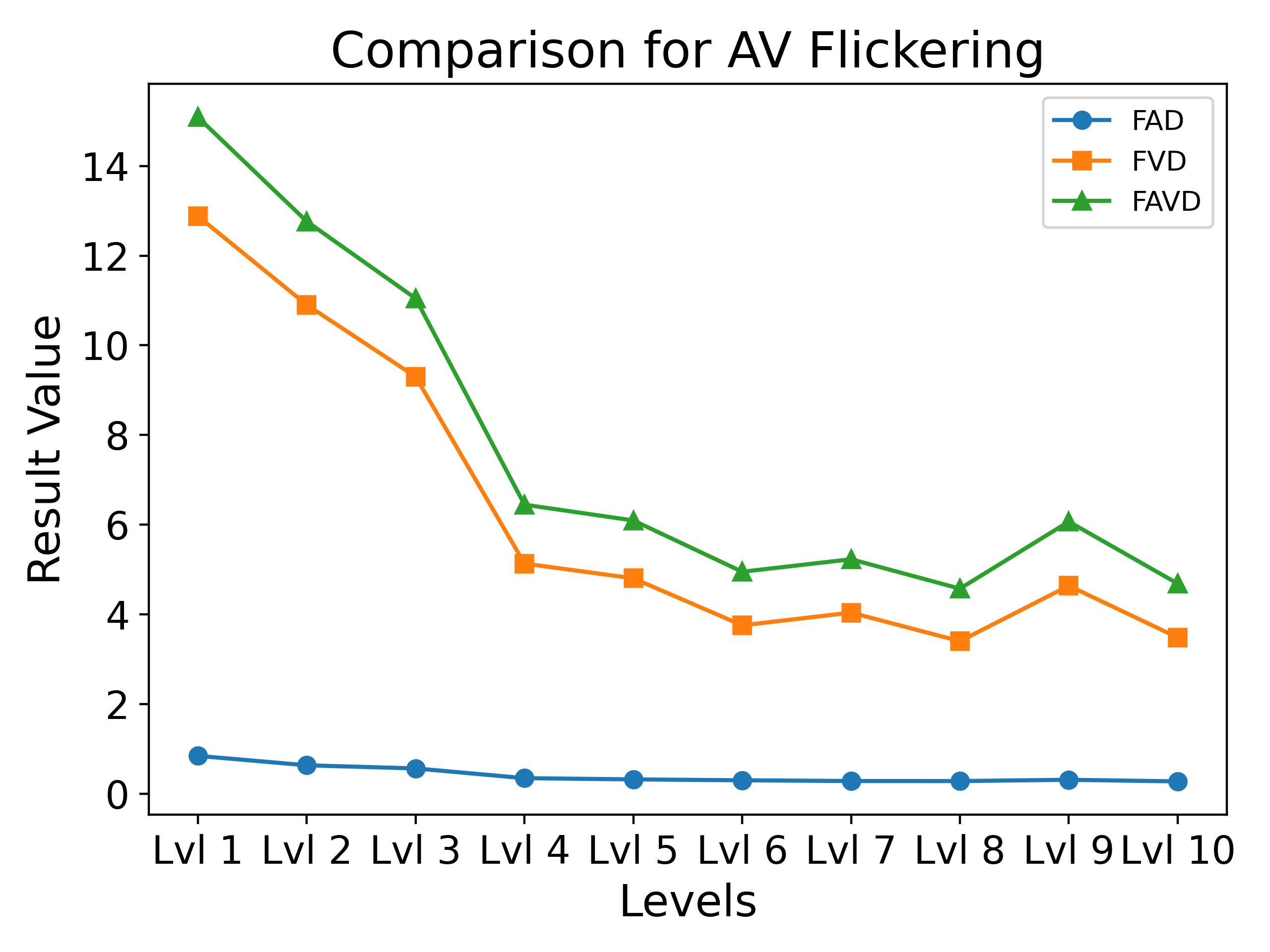}
        \subcaption{AV Flickering}
    \end{minipage}

    \caption{In these plots, we show the effect of distortions with varying levels on \textcolor{orange}{\textbf{FAD}}, \textcolor{blue}{\textbf{FVD}}, and \textcolor{green}{\textbf{FAVD}}. Flat trend line implies that a metric is not able to capture the distortion type. Increasing or decreasing trends show that a metric is susceptible to varying levels in a distortion. Distortion levels are taken from Table~\ref{tab:type_values}.}
    \label{fig:fd_metrics_detailed}
\end{figure}

\section{PEAVS v/s SparseSync Benchmark} \label{sec:sparsesyncvsavsync} 
In our comparison for PEAVS v/s SparseSync, we created a held out evaluation set specifically for the head to head comparison. In addition to the 37 videos (annotated from AVS benchmark), we selected 163 more diverse set of videos from AudioSet \cite{audioset}, ensuring that they do not contain talking faces but showcase various real-world scenarios, such as a car driving by, an instrument being played, a dog barking, and many more. We manually looked into each video to ensure if they are well aligned. We then changed the offset of audio stream of these 200 videos by randomly sampling from the different levels of audio-shift distortion (as described in Table~\ref{tab:type_values}). This resulted in a total of 400 videos with an equal mix of ground truth and distorted videos. This portion of evaluation set will also be released as a part of AVS benchmark. 

\paragraph{Levels of Distortion}
ITU recommends an acceptability thresholds for AV shift up to 185 ms~\cite{noauthor_bt1359_nodate}, this is why we consider distortions of 0.045, 0.1 and $\pm$0.125 as positive cases (i.e. ground truth). This also aligns well with SparseSync buckets where the first distortion starts at $\pm$0.2 seconds (i.e. SparseSync's step-size).





\end{document}